\documentclass{bmvc2k}

\usepackage{times}
\usepackage{epsfig}
\usepackage{graphicx}
\usepackage{amsmath}
\usepackage{amssymb}
\usepackage{enumitem}
\usepackage{algorithm}
\usepackage[noend]{algpseudocode}
\usepackage{subfigure}

\title{Multi-label Zero-shot Classification by Learning to Transfer from External Knowledge}

\addauthor{He Huang}{hehuang@uic.edu}{1}
\addauthor{Yuanwei Chen}{liaoyuan.zhangfei@alibaba-inc.com}{2}
\addauthor{Wei Tang}{tangw@uic.edu}{1}
\addauthor{Wenhao Zheng}{zwh149850@alibaba-inc.com}{2}
\addauthor{Qing-Guo Chen}{qingguo.cqg@alibaba-inc.com}{2}
\addauthor{Yao Hu}{yaoohu@alibaba-inc.com}{2}
\addauthor{Philip Yu}{psyu@uic.edu}{1}

\addinstitution{
 Department of Computer Science,\\
 University of Illinois at Chicago,\\
 Chicago, USA
}
\addinstitution{
 Alibaba Group,\\
 Hangzhou, China
}

\runninghead{Arxiv submission}{Multi-label Zero-shot Classification}

\def\eg{\emph{e.g}\bmvaOneDot}

\def\etal{\emph{et al}\bmvaOneDot}

\newcommand{\ie}{\emph{i.e.}} 

\begin{document}

\maketitle

\begin{abstract}
Multi-label zero-shot classification aims to predict multiple unseen class labels for an input image. It is more challenging than its single-label counterpart. On one hand, the unconstrained number of labels assigned to each image makes the model more easily overfit to those seen classes. On the other hand, there is a large semantic gap between seen and unseen classes in the existing multi-label classification datasets. To address these difficult issues, this paper introduces a novel multi-label zero-shot classification framework by learning to transfer from external knowledge. We observe that ImageNet is commonly used to pretrain the feature extractor and has a large and fine-grained label space. This motivates us to exploit it as external knowledge to bridge the seen and unseen classes and promote generalization. Specifically, we construct a knowledge graph including not only classes from the target dataset but also those from ImageNet. Since ImageNet labels are not available in the target dataset, we propose a novel PosVAE module to infer their initial states in the extended knowledge graph. Then we design a relational graph convolutional network (RGCN) to propagate information among classes and achieve knowledge transfer. Experimental results on two benchmark datasets demonstrate the effectiveness of the proposed approach.
\end{abstract}
\vspace{-2em}

\section{Introduction}
\vspace{-1em}

Recent deep learning methods are able to outperform humans in image classification tasks by leveraging supervised learning and large-scale datasets such as ImageNet~\cite{ILSVRC15ImageNet}. However, machines still cannot compete with humans when the task requires generalizing learned concepts to novel concepts. For example, a person who has only seen horses but not zebras can still recognize a zebra if he/she is told that ``a zebra is like a horse but has black and white strips''. 
To evaluate a model's ability to generalize what it has learned during training to totally new concepts in testing, the task of zero-shot learning (ZSL)~\cite{xian2018gbu,akata2013ALE,GDAN_2019_CVPR, schonfeld2019generalized,chen2019hybrid} divides a set of classes into two disjoint sets: a set of \textit{seen} classes and a set of \textit{unseen} classes. Images from \textit{seen} classes are provided during training, while images of both \textit{seen} and \textit{unseen} classes are used for testing. 

\begin{figure}[!ht]
\centering
\includegraphics[width=0.5\linewidth]{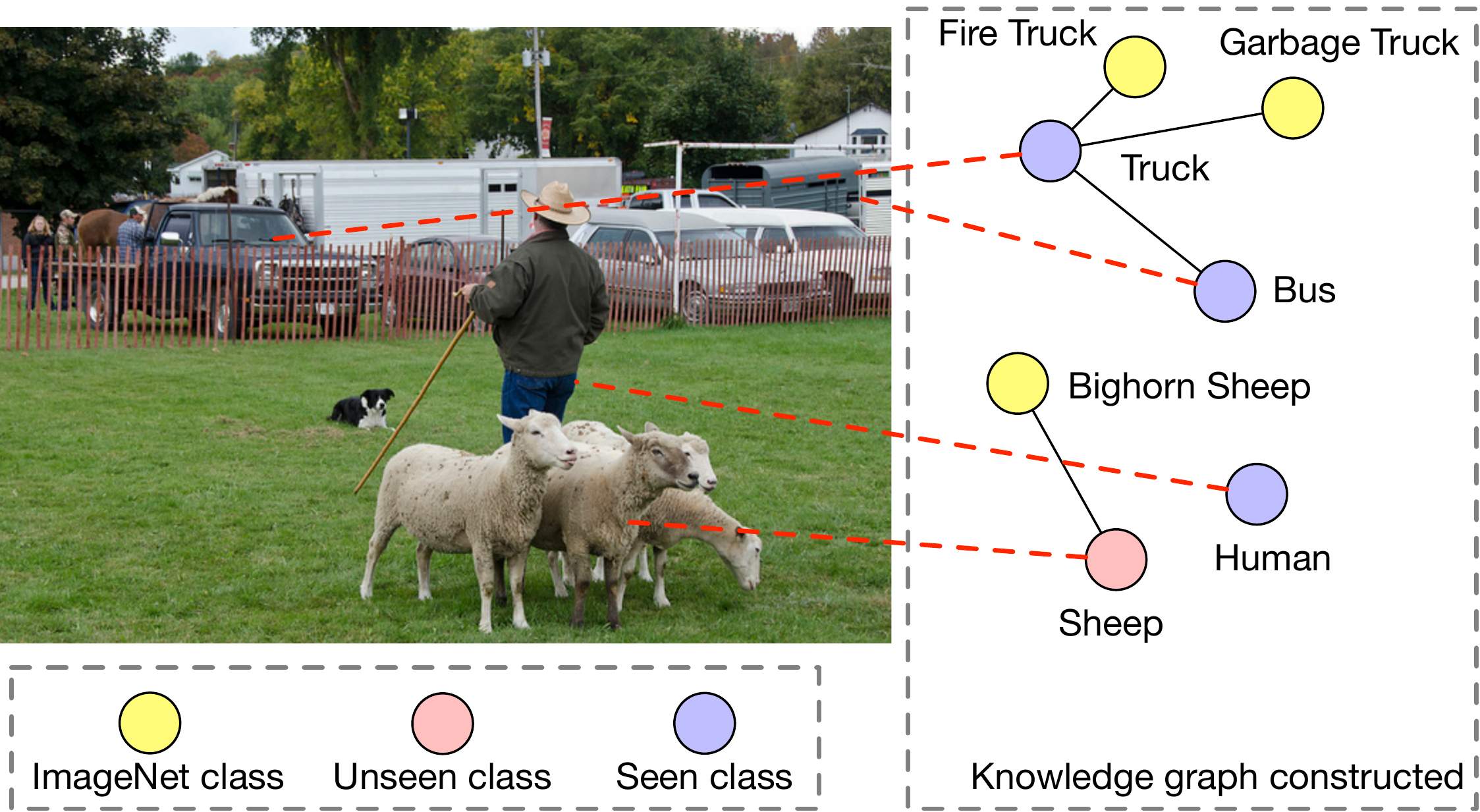}
\caption{Illustration of combining ImageNet classes into multi-label zero-shot image classification on MS-COCO~\cite{lin2014mscoco}. Without the yellow nodes from ImageNet~\cite{ILSVRC15ImageNet}, it would be hard to predict the unseen class \textit{Sheep} since other seen classes are not semantically related to it. However, with the knowledge that the current image has a high probability on the ImageNet class \textit{Bighorn Sheep} and the word embeddings of both \textit{Sheep} and \textit{Bighorn Sheep}, the model can better predict the unseen class by referring to the ImageNet classes. 
}
\vspace{-1em}
\label{fig:moti}
\end{figure}


Although zero-shot learning for single-label classification has been well studied~\cite{xie2019attentive,mishra2017cvae-zsl, xian2018fganzsl, verma2018segzsl, GDAN_2019_CVPR, schonfeld2019generalized}, multi-label zero-shot learning (ML-ZSL)~\cite{mensink2014costa,fu2015transductive,zhang2016fast,lee2018multi} is a less explored area. An image is associated with potentially multiple seen and unseen classes, but only labels of seen classes are provided during training. The multi-label zero-shot learning problem is more difficult than its single-label counterpart for two reasons. First, since there is no constraint on the number of labels assigned to an image and only labels of seen classes are provided during training, the model can easily bias towards the seen classes and ignore the unseen ones. Second, the datasets used for multi-label zero-shot learning are generally much more challenging than those used for traditional zero-shot learning. While the datasets used for zero-shot learning (ZSL) usually consist of closely related classes such as different kinds of birds (\textit{e.g.,} \textit{Baird Sparrow} and \textit{Chipping Sparrow} in CUB~\cite{wah2011cub_bird}), the datasets for multi-label classification contain high-level concepts that are less related to each other (\textit{e.g.,} \textit{truck} and \textit{sheep} in MS-COCO~\cite{lin2014mscoco}). The semantic gap between seen and unseen classes in multi-label datasets makes it difficult for the model to generalize knowledge learned from seen classes to predict unseen classes.

Some studies~\cite{lee2018multi,zhang2016fast,fu2015transductive,gaure2017probabilistic,marino2016more,ren2017multiple,mensink2014costa} try to solve the multi-label zero-shot learning problem by leveraging word embeddings~\cite{zhang2016fast} or constructing structured knowledge graphs~\cite{lee2018multi}. Although these methods empirically work well, none of them utilizes external knowledge to close the gap between semantically distinct seen and unseen classes. By contrast, humans learn new concepts by connecting them with those that they have already learned before, and thus can transfer knowledge from previous experience to solve new tasks. This ability is called \textit{transfer learning}. The most popular way to use transfer learning in various computer vision tasks is to extract image features from classifiers (\textit{e.g.,} ResNet~\cite{he2016resnet}) pretrained on ImageNet~\cite{ILSVRC15ImageNet}. However, this naive approach does not make use of the semantic information of the 1K ImageNet classes, which can be helpful in recognizing unseen classes in zero-shot learning, especially when none of the seen classes is closely related to the unseen ones. 

In this paper, we propose to build a knowledge graph with not only the classes from the target multi-label dataset (\textit{e.g., MS-COCO~\cite{lin2014mscoco}}) but also classes from the external single-label dataset (\textit{e.g., ImageNet~\cite{ILSVRC15ImageNet}}). Then classification can be performed via graph reasoning methods such as graph convolutional networks (GCNs)~\cite{kipf2016gcn}. 
Our intuition is that the knowledge of which ImageNet classes are more related to the image can help close the semantic gap between the seen and unseen classes.
For example, as illustrated in Figure~\ref{fig:moti}, the unseen class \textit{Sheep} in the image is hard to predict since it is semantically very different from all seen classes in the dataset. But if we include the 1K ImageNet classes into the knowledge graph, there will be a class \textit{Bighorn Sheep} which is semantically related to \textit{Sheep}. Given the predicted probability distribution on the ImageNet classes, if the \textit{Bighorn Sheep} class has a high probability, we can infer that the image contains the \textit{Sheep} class.

After incorporating the ImageNet classes as nodes in the knowledge graph of the target dataset, it is still nontrivial to infer their states. A simple solution is to use the predicted probability distribution on ImageNet classes as pseudo labels, and then we can treat the ImageNet classes the same as those in the target multi-label dataset. However, this simple approach is problematic. Since the ImageNet classes are not labeled in the multi-label dataset, treating them the same as the real target labels may confuse the network and lead to inferior learning. 
To address this issue, we propose a novel PosVAE module based on a  conditional variational auto-encoder~\cite{sohn2015cvae} to model the posterior probabilities of the ImageNet classes. Then the states of ImageNet nodes are obtained via inference through the PosVAE network.
The states of nodes from the target dataset are obtained by another semantically conditioned encoder which takes as input the concatenation of the image features and the class embedding of each node. The states of all nodes are passed into a relational graph convolutional network (RGCN), which propagates information among classes and generates predictions for each class in the target dataset.

Our contributions are summarized as follows:
\begin{itemize}[leftmargin=*,noitemsep,topsep=0pt]
    \item To the best of our knowledge, this is the first attempt to transfer the semantic knowledge from the 1K ImageNet classes to help solve the multi-label zero-shot learning problem. This is achieved by adding the ImageNet classes as additional nodes to extend the knowledge graph constructed by labels in the target multi-label dataset. By contrast, previous methods only utilize pretrained ImageNet classifiers as feature extractors.
    \item We design a PosVAE module, which is based on a conditional variational auto-encoder~\cite{sohn2015cvae}, to infer the  states of ImageNet nodes in the extended knowledge graph, and a relational graph convolutional network (GCN) to generate predictions for seen and unseen classes.
    \item We conduct extensive experiments and demonstrate that the proposed approach can effectively transfer knowledge from the ImageNet classes and improve the multi-label zero-shot classification performance. We also show that it outperforms previous methods.
\end{itemize}
\vspace{-1em}

\section{Related Work}
\vspace{-1em}
\textbf{Multi-label Classification.} 
Recently, some work~\cite{wang2016cnnrnn,wang2017multi,zhu2017spatialreg,chen2019mlgcn,chen2019ssgrl,marino2016more} propose different ways to learn the correlations among labels. Marino~\etal\cite{marino2016more} first detect one label and then sequentially predict more related labels of the input image. Wang~\etal\cite{wang2016cnnrnn} use a recurrent neural network (RNN) to capture the dependencies between labels, while Zhu~\etal\cite{zhu2017spatialreg} utilize spatial attention to better capture spatial and semantic correlations of labels. Chen~\etal\cite{chen2019mlgcn} construct a knowledge graph from labels and use the co-occurrence probabilities of labels as weights of the edges in the graph, and apply a graph convolutional network (GCN)~\cite{kipf2016gcn} to perform multi-label classification. Chen~\etal\cite{chen2019ssgrl} use the pretrained word embeddings of labels as semantic queries on the image feature map and apply spatial attention to obtain semantically attended features for each label. Although these methods consider label dependencies and work well on traditional multi-label classification, they are not directly applicable in the multi-label zero-shot  setting because the label co-occurrence statistics is not available for unseen classes, and only utilizing the statistics for seen labels can cause severe bias towards seen classes. Thus we need a different way to learn the relations between labels in the multi-label zero-shot setting.


\textbf{Multi-label Zero-shot Learning} is a less explored problem compared to single-label zero-shot learning~\cite{mishra2017cvae-zsl, xian2018fganzsl, verma2018segzsl, GDAN_2019_CVPR, schonfeld2019generalized}. Fu~\etal\cite{fu2015transductive} enumerate all possible combinations of labels and thus transform the problem into single-label zero-shot learning. However, as the number of labels increases, the number of label combinations will increase exponentially, which makes this method inapplicable. Mensink~\etal\cite{mensink2014costa} learn classifiers for unseen classes as weighted combinations of seen classifiers learned with co-occurrence statistics of seen labels. Zhang \etal\cite{zhang2016fast} consider the projected image features as a projection vector which ranks relevant labels higher than irrelevant ones by calculating the inner product between the projected image features and label word embeddings. Gaure~\etal\cite{gaure2017probabilistic} further assume that the co-occurrence statistics of both seen and unseen labels are available and design a probabilistic model to solve the problem. Ren~\etal\cite{ren2017multiple} cast the problem into a multi-instance learning~\cite{zha2008joint} framework and learn a joint latent space for both image features and label embeddings. Lee~\etal\cite{lee2018multi} construct a knowledge graph for both seen and unseen labels, and learn a gated graph neural network (GGNN)~\cite{li2015ggnn} to propagate predictions among labels. To our knowledge, none of the previous work considers the large semantic gap between seen and unseen labels, 
and we are the first to exploit the label space of a single-label image classification dataset as external knowledge and transfer it for zero-shot multi-label image classification.
\vspace{-2em}

\section{Approach}
\vspace{-0.5em}
\subsection{Problem Definition and Notations}
\vspace{-0.5em}
The multi-label zero-shot classification problem is defined as follows. Given a multi-label classification dataset, the label space $\mathcal{Y}$ is split into seen labels $\mathcal{Y}^s$ and unseen labels $\mathcal{Y}^u$ such that $\mathcal{Y}=\mathcal{Y}^s\cup \mathcal{Y}^u$ and $\mathcal{Y}^s\cap \mathcal{Y}^u=\emptyset$. During training, a set of image-label pairs is available, where the labels $\mathbf{y}$ of an image $\mathbf{x}$ come from only the seen classes: $\{\mathbf{x},\mathbf{y}|\mathbf{y} \subseteq \mathcal{Y}^s\}$. During testing, the images may have labels from both seen and unseen classes: $\{\mathbf{x},\mathbf{y}|\mathbf{y} \subseteq \mathcal{Y}\}$. The goal is to learn a model $f:\mathbf{x}\rightarrow \mathcal{Y}^s\cup\mathcal{Y}^u$, given only seen classes $\mathcal{Y}^s$ during training. In this work, we propose to use an external set of labels $\mathcal{Y}^a$ from ImageNet~\cite{ILSVRC15ImageNet} where $\mathcal{Y}^a\cap\mathcal{Y}=\emptyset$. For an arbitrary class $y$, its pretrained GloVe~\cite{pennington2014glove} embedding is denoted by $\textbf{s}_y$, and all class embeddings are aggregated into a single matrix $\mathbf{S}\in \mathbb{R}^{(|\mathcal{Y}|+|\mathcal{Y}^a|)\times d_s}$, where $d_s$ is the dimension of embeddings.
\vspace{-0.5em}

\begin{figure*}[t]
\centering
\includegraphics[width=0.9\linewidth]{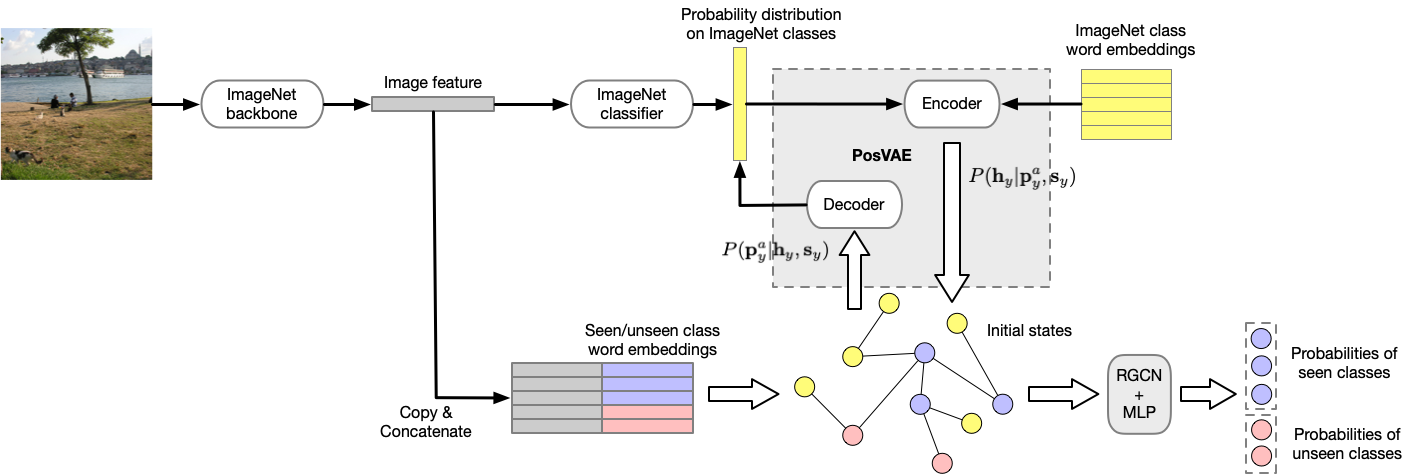}
\caption{Pipeline of the proposed model. 
We hide the self-connections here for clarity.
}
\vspace{-1.5em}
\label{fig:model}
\end{figure*}
\vspace{-0.5em}
\subsection{Overview of the Proposed Method}
\vspace{-0.5em}
The overall pipeline of the proposed model is illustrated in Figure~\ref{fig:model}. An input image is first fed into a pretrained ImageNet classifier to extract its image feature and a probability distribution $\mathbf{p}^a$ over the ImageNet classes $\mathcal{Y}^a$. We first construct a graph whose nodes are classes from the target dataset.
We calculate the pair-wise Wu-Palmer (WUP)~\cite{wu1994wup} similarities on WordNet~\cite{miller1995wordnet} and then use a threshold to determine whether an edge exists between a pair of nodes. Self-connections are also included. During training, only seen classes  $\mathcal{Y}^s$ are included into the training graph, while during testing both seen and unseen classes ($\mathcal{Y}=\mathcal{Y}^s\cup\mathcal{Y}^u$) are included in the testing graph. The classes $\mathcal{Y}^a$ from ImageNet~\cite{ILSVRC15ImageNet} are also added to the training and testing graphs, and edges are added in the same way as described above. The overlapping classes between ImageNet and the target dataset have been removed from $\mathcal{Y}^a$.
All aggregated edges form an adjacency matrix $\mathbf{A}\in \mathbb{R}^{(|\mathcal{Y}^s|+|\mathcal{Y}^a|)\times(|\mathcal{Y}^s|+|\mathcal{Y}^a|)}$ for training, and an adjacency matrix $\mathbf{A}\in \mathbb{R}^{(|\mathcal{Y}|+|\mathcal{Y}^a|)\times(|\mathcal{Y}|+|\mathcal{Y}^a|)}$ for testing (with little abuse of notation).

Since the target dataset does not have ground-truth labels of the ImageNet classes $\mathcal{Y}^a$, we need to treat $\mathcal{Y}$ and $\mathcal{Y}^a$ differently when inferring their states in the graph. For nodes in $\mathcal{Y}^s\cup\mathcal{Y}^u$, we infer their initial states by concatenating their corresponding word embeddings and the extracted image features. Then a multi-layer perceptron (MLP) network is applied to obtain the node features $\mathbf{H}^d\in \mathbb{R}^{|\mathcal{Y}^s|\times d_h}$ for training and $\mathbf{H}^d\in \mathbb{R}^{|\mathcal{Y}|\times d_h}$ for testing, where $d_h$ is the dimension of node features. Meanwhile, for nodes corresponding to $\mathcal{Y}^a$, we design a PosVAE module to estimate the posterior $P(\mathbf{h}_y|\mathbf{p}^a_y, \mathbf{s}_y)$ for each $y\in\mathcal{Y}^a$, where $\mathbf{h}_y$ is the node state of class $y$ and $\mathbf{p}^a_y$ is predicted probability of the class $y\in\mathcal{Y}^a$. Then we sample from this posterior distribution to infer the initial node states for all classes in $\mathcal{Y}^a$ denoted as $\mathbf{H}^a \in \mathbb{R}^{|\mathcal{Y}^a|\times d_h}$. We will discuss the detailed design of PosVAE in Section~\ref{sec:vae}. Then we can obtain the initial node states for all the nodes in the graph by concatenation: $\mathbf{{H}}\in \mathbb{R}^{(|\mathcal{Y}^s|+|\mathcal{Y}^a|)\times d_h}$ for training and $\mathbf{{H}}\in \mathbb{R}^{(|\mathcal{Y}|+|\mathcal{Y}^a|)\times d_h}$  for testing.

With the initial node states $\mathbf{{H}}$, the adjacency matrix $\mathbf{A}$, and the embedding matrix $\mathbf{S}$, we design a Relational Graph Convolutional Network (RGCN) to propagate information among the nodes and refine their features, which will be discussed in Section~\ref{sec:rgcn}. The output features of RGCN are fed into another MLP network followed by Sigmoid functions to predict the existence of each label in the target dataset ($\mathcal{Y}^s$ for training  and $\mathcal{Y}$ for testing). 
\vspace{-1em}

\subsection{Variational Inference for Node States of $\mathcal{Y}^a$}
\label{sec:vae}
\vspace{-0.5em}

For each ImageNet class $y$, we aim to estimate its corresponding initial node state $\mathbf{h}_y$ in the graph from its probability $\mathbf{p}^a_y$ predicted by pretrained ImageNet classifier and its embedding $\mathbf{s}_y$. 
The objective is to estimate the posterior $P(\mathbf{h}_y|\mathbf{p}^a_y, \mathbf{s}_y)$. Since we do not have the exact form of the posterior, we resort to variational inference and use another learned distribution $Q(\mathbf{h}_y|\mathbf{p}^a_y, \mathbf{s}_y)$ to approximate the true posterior $P(\mathbf{h}_y|\mathbf{p}^a_y, \mathbf{s}_y)$. In order to ensure a good approximation, we aim to minimize the KL-divergence between $Q(\mathbf{h}_y|\mathbf{p}^a_y, \mathbf{s}_y)$ and $P(\mathbf{h}_y|\mathbf{p}^a_y, \mathbf{s}_y)$:
\begin{align}
    \mathcal{L}_{VAE} & = \frac{1}{|\mathcal{Y}^a|} \sum_{y \in \mathcal{Y}^a} D_{KL}[Q(\mathbf{h}_y|\mathbf{p}^a_y, \mathbf{s}_y) || P(\mathbf{h}_y|\mathbf{p}^a_y, \mathbf{s}_y)] \\
             & = \frac{1}{|\mathcal{Y}^a|} \sum_{y \in \mathcal{Y}^a} \mathbb{E}_{\mathbf{h}_y\sim  P(\mathbf{h}_y|\mathbf{p}^a_y, \mathbf{s}_y)} [\log Q(\mathbf{h}_y|\mathbf{p}^a_y, \mathbf{s}_y) - \log{ P(\mathbf{h}_y|\mathbf{p}^a_y, \mathbf{s}_y)}]
\end{align}

Similar to the conditional variational auto-encoder~\cite{sohn2015cvae}, the above objective can be further derived as:
\begin{align}
    \mathcal{L}_{VAE} & = -\frac{1}{|\mathcal{Y}^a|} \sum_{y \in \mathcal{Y}^a} [\mathbb{E}_{\mathbf{h}_y\sim P(\mathbf{h}_y|\mathbf{p}^a_y, \mathbf{s}_y)}  [\log{P(\mathbf{p}^a_y|\mathbf{h}_y,\mathbf{s}_y)} - \log Q(\mathbf{h}_y|\mathbf{p}^a_y, \mathbf{s}_y) + \log{P(\mathbf{h}_y)}] - P(\mathbf{p}^a_y)]\\
    & =  -\frac{1}{|\mathcal{Y}^a|} \sum_{y \in \mathcal{Y}^a}[\mathbb{E}_{\mathbf{h}_y\sim P(\mathbf{h}_y|\mathbf{p}^a_y, \mathbf{s}_y)}[\log{P(\mathbf{p}^a_y|\mathbf{h}_y,\mathbf{s}_y)}] - D_{KL}[Q(\mathbf{h}_y|\mathbf{p}^a_y, \mathbf{s}_y) || P(\mathbf{h}_y)]],
    \label{eq:vae}
\end{align}
where the term $P(\mathbf{p}^a_y)$ is neglected since it does not depend on $\mathbf{h}_y$ during optimization.


We assume $\mathbf{h}_y$ follows an isotropic Gaussian distribution $Q(\mathbf{h}_y|\mathbf{p}^a_y, \mathbf{s}_y) \sim \mathcal{N}(\mu,\sigma^2\mathbf{I})$ and $P(\mathbf{h}_y)\sim \mathcal{N}(0,1)$. By using the reparameterization trick~\cite{kingma2013vae}, the initial node states $\mathbf{h}_y$ can be derived as:
\begin{align}
    \mathbf{z} \sim \mathcal{N}(0,1);\  
    (\mu, \log\sigma) = \textrm{Encoder}(\mathbf{p}^a_y, \mathbf{s}_y);\ 
    \mathbf{h}_y = \mu + \exp(\log\sigma/2)\odot\mathbf{z},
\end{align}
where the Encoder network is a stack of MLP layers and $\odot$ is element-wise multiplication. The Encoder first projects $\mathbf{p}^a_y$ into the same dimension space as $\mathbf{s}_y$. Then the two vectors are concatenated and fed into subsequent MLP layers to generate the mean $\mu \in\mathbb{R}^{d_h}$ and log standard deviation $\log\sigma \in\mathbb{R}^{d_h}$. 
In order to calculate the log-likelihood term $\log{P(\mathbf{p}^a_y|\mathbf{h}_y,\mathbf{s}_y)}$, we assume a Gaussian distribution on $\mathbf{p}^a_y$, and use the Decoder network to project the concatenation of $\mathbf{h}_y$ and $\mathbf{s}_y$ back to $\mathbf{p}^a_y$. Then maximizing the log-likelihood term is approximately equal to minimizing the mean-square error (MSE) between the reconstructed probability and the real one.
\vspace{-1em}

\subsection{Relational Graph Convolutional Network}
\vspace{-0.5em}

\label{sec:rgcn}
The Graph Convolutional Network (GCN)~\cite{kipf2016gcn} was first introduced as a method to perform semi-supervised classification. Its core idea is to design a message passing mechanism for graph data so that the node features can be updated recursively and become
more suitable to the desired task. Different from the conventional convolution that operates in the Euclidean space like images, a GCN works on graph-structure data that contain nodes and edges but without fixed geometric layout. Given the  states $\mathbf{H}^l$ of the current layer $l$ and the adjacency matrix $\mathbf{A}$, the states of the $(l+1)$-th layer can be calculated as:
\begin{align}
    \mathbf{H}^{l+1} = \delta(\hat{\mathbf{A}}\mathbf{H}^l\mathbf{W}^l),
\end{align}
where $\mathbf{W}^l$ is the learnable weight matrix of the $l$-th layer, $\hat{\mathbf{A}}$ is the normalized version of $\mathbf{A}$ \cite{kipf2016gcn}, and $\delta(\cdot)$ is a non-linear activation function such as ReLU.

In traditional multi-label classification models, the $\mathbf{A}$ matrix is often represented by the co-occurrence probabilities between labels~\cite{chen2019ssgrl, chen2019mlgcn}. However, in the zero-shot learning setting, we do not have this information. Thus we set $\mathbf{A}\in \mathbb{R}^{(|\mathcal{Y}^s|+|\mathcal{Y}^a|)\times(|\mathcal{Y}^s|+|\mathcal{Y}^a|)}$ as binary adjacency matrix and learn a new matrix $\hat{\mathbf{W}}\in \mathbb{R}^{(|\mathcal{Y}^s|+|\mathcal{Y}^a|)\times(|\mathcal{Y}^s|+|\mathcal{Y}^a|)}$ to indicate the weight of each edge ( $\mathbf{A},\mathbf{\hat{W}}\in \mathbb{R}^{(|\mathcal{Y}|+|\mathcal{Y}^a|)\times(|\mathcal{Y}|+|\mathcal{Y}^a|)}$ for testing). In order to learn the weight matrix, we propose the relational graph convolutional network (RGCN) as follows. Specifically, the weight of the edge between a pair of classes $(u,v)$ is calculated by first concatenating their word embeddings $(\mathbf{s}_u,\mathbf{s}_v)$ together and feeding them into an MLP network to generate a scalar which represents the weight. In this way, we are able to calculate the weight between seen and unseen classes. An entry $w_{uv}$ of the weight matrix $\mathbf{\hat{W}}$ is derived by:
\begin{align}
    w_{uv} = \textrm{MLP}(\mathbf{s}_u, \mathbf{s}_v),
\end{align}
where $w_{uv}\in [0,1]$, and $(\mathbf{s}_u, \mathbf{s}_v) \in \mathbb{R}^{d_s}$. Elements on the diagonal of $\mathbf{W}$ are set to 1 to include self-connections. After obtaining the weight matrix $\mathbf{\hat{W}}$, we calculate $\hat{\mathbf{A}}=\textrm{normalize}(\mathbf{\hat{W}}\odot\mathbf{A})$, where $\odot$ is element-wise multiplication. After obtaining the new matrix $\hat{\mathbf{A}}$, the rest of RGCN is the same as the traditional GCN~\cite{kipf2016gcn}. After obtaining the node states $\mathbf{H}^s \in\mathbb{R}^{|\mathcal{Y}^s|\times d_h}$ of each node corresponding to the seen classes $\mathcal{Y}^s$, we apply a final MLP layer to the node states to calculate the probability of each label, \ie, $\mathbf{p}^s = \textrm{MLP}(\mathbf{H}^s), \mathbf{p}^s\in\mathbb{R}^{|\mathcal{Y}^s|}$, and $L$ is the last layer of RGCN. We train the relational graph convolutional network (RGCN) with class-wise binary cross-entropy loss:
\begin{align}
    \mathcal{L}_{RGCN} = -\frac{1}{|\mathcal{Y}^s|}\sum_{y \in \mathcal{Y}^s} [(1-\mathbf{I}(y))\log (1-\mathbf{p}^s_y) + \mathbf{I}(y)\log \mathbf{p}^s_y],
\end{align}
where $\mathbf{I}(y)$ is an indicator function that outputs 1 if the input image contains label $y$, otherwise 0.
The whole model is trained by minimizing the joint loss function:
\begin{align}
    \mathcal{L}_{all} = \mathcal{L}_{RGCN} + \lambda\mathcal{L}_{VAE},
\end{align}
where $\lambda$ is a hyper-parameter to balance the two terms.
\vspace{-1em}
\section{Experiments}
\vspace{-0.5em}
\subsection{Datasets and Setting}
\vspace{-0.5em}

\textbf{Datasets}. In our experiments, we use two multi-label datasets, \ie, MS-COCO~\cite{lin2014mscoco} and NUS-WIDE~\cite{chua2009nuswide}. MS-COCO~\cite{lin2014mscoco} is a large-scale dataset commonly used for multi-label image classification, object detection, instance segmentation and image captioning. We adopt the 2014 challenge which contains 79,465 training images and 40,137 testing images respectively (after removing images without labels). Among its 80 classes, we randomly select 16 unseen classes and make sure that the unseen classes do not overlap with the ImageNet classes~\cite{ILSVRC15ImageNet}. The 16 unseen classes are \textit{('bicycle', 'boat', 'stop sign', 'bird', 'backpack', 'frisbee', 'snowboard', 'surfboard', 'cup', 'fork', 'spoon', 'broccoli', 'chair', 'keyboard', 'microwave', 'vase')}. NUS-WIDE~\cite{chua2009nuswide} is a web-crawled dataset with 54,334 training images and 42,486 testing images (after removing images without labels). Among the 81 classes of NUS-WIDE~\cite{chua2009nuswide}, we randomly select 16 unseen classes and make sure that the unseen classes do not overlap with the ImageNet classes~\cite{ILSVRC15ImageNet}. The 16 unseen classes are \textit{('airport', 'cars', 'food', 'fox', 'frost', 'garden', 'mountain', 'police', 'protest', 'rainbow', 'sun', 'tattoo', 'train', 'water', 'waterfall', 'window')}.

\begin{table*}[!t]
\centering
\caption{Performance comparison on MS-COCO~\cite{lin2014mscoco} and NUS-WIDE~\cite{chua2009nuswide}}.
\begin{tabular}{|l|c|c|c|c|c|c|c|c|}
\hline
 & \multicolumn{4}{c|}{MS-COCO} & \multicolumn{4}{c|}{NUS-WIDE} \\ \hline
 & \multicolumn{2}{c|}{Seen} & \multicolumn{2}{c|}{Unseen} & \multicolumn{2}{c|}{Seen} & \multicolumn{2}{c|}{Unseen} \\ \hline
Model & mAP & miAP & mAP & miAP & mAP & miAP & mAP & miAP \\ \hline
visual-semantic & 58.30 & 59.87 & 18.60 & 28.08 & \textbf{25.79} & 34.64 & \textbf{12.73} & 29.01 \\ \hline
fast0tag & 44.60 & 60.20 & 19.10 & 32.90 & 25.10 & \textbf{36.09} & 10.41 & 34.58 \\ \hline
SKG & \textbf{64.02} & \textbf{64.22} & 18.03 & 33.68 & 22.89 & 32.95 & 8.89 & 33.77 \\ \hline
RGCN & 53.67 & 60.43 & 18.97 & 33.65 & 24.95 & 31.23 & 9.98 & 33.61 \\ \hline
RGCN-XL & 34.10 & 49.18 & 17.32 & 35.43 & 23.66 & 30.95 & 9.86 & 34.31 \\ \hline
RGCN-PosVAE & 51.54 & 57.13 & \textbf{20.52} & \textbf{37.14} & 24.67 & 32.09 & 11.09 & \textbf{36.78} \\ \hline
\end{tabular}
\vspace{-1.5em}
\label{tab:result}
\end{table*}

\textbf{Graph construction}. We calculate the WUP~\cite{wu1994wup} similarity between each pair of classes, and an edge is added to the graph if the corresponding WUP similarity is greater than a specified threshold (0.5 in all of our experiments). When adding the ImageNet~\cite{ILSVRC15ImageNet} classes into either the MS-COCO~\cite{lin2014mscoco} or NUS-WIDE~\cite{chua2009nuswide} graph, we exclude any class existing in the target dataset from ImageNet to ensure the additional nodes are distinct from the nodes in the two target datasets.

\textbf{Evaluation protocol}. We adopt two evaluation metrics, \ie, \textit{mean average precision} (\textbf{mAP}) which evaluates the model performance on the class level, and \textit{mean image average precision} (\textbf{miAP}) which evaluates the model performance on the image level. We calculate the \textbf{mAP} and \textbf{miAP} for two sets of labels, \ie, \textbf{seen} classes $\mathcal{Y}^s$ and \textbf{unseen} classes  $\mathcal{Y}^u$.

\textbf{Baselines}. We use three models from previous work and some variants of our proposed method as baselines. \textbf{Visual-semantic}~\cite{akata2013ALE} was proposed for single-label zero-shot classification, which projects the visual features of images to the semantic space of classes and use the class embeddings as classifiers. Here we change the loss function from multi-class cross-entropy to binary cross-entropy applied to each class individually. \textbf{Fast0Tag}~\cite{zhang2016fast} uses class embeddings as projection vectors to rank the closeness between images and classes, and applies a triplet-based ranking loss to train the model. \textbf{SKG}~\cite{lee2018multi} constructs a structural knowledge graph from labels and applies GGNN~\cite{li2015ggnn} to infer the existence of each label. It also uses WUP simiarities on WordNet~\cite{miller1995wordnet} to construct the graph as we do. \textbf{RGCN} is a basic version of our proposed model which does not include external nodes from ImageNet~\cite{ILSVRC15ImageNet}. \textbf{RGCN-XL} is another variant of our proposed model, where the ImageNet~\cite{ILSVRC15ImageNet} nodes are also included in the graph but treated the same as the other nodes in the target dataset. For ImageNet nodes, we predict their probabilities in the same way as the other classes in the target dataset, but train them with MSE to reconstruct the probability distribution of the pretrained ImageNet classifier. \textbf{RGCN-PosVAE} is our full model which uses PosVAE to infer the hidden states of ImageNet classes and applies RGCN to propagate information among different classes.

\textbf{Implementation details}.
For SKG~\cite{lee2018multi} we use the code provided by the authors and modify it to fit our pipeline, and we implement all other baselines. We use ResNet-101~\cite{he2016resnet} pretrained on ImageNet~\cite{ILSVRC15ImageNet} as the feature extractor. We use the 300-dimension pretrained GloVe~\cite{pennington2014glove} vectors to represent class embeddings. We implement the encoder and decoder of PosVAE as MLP networks with a hidden layer of size 256, while the relation network that computes $w_{uv}$ is implemented as a two-layer MLP network with a hidden size 256. We use two layers of RGCN with the feature dimension 256. We use an Adam~\cite{kingma2014adam} optimizer and set the initial learning rate as $5e^{-4}$ and decay it by 0.1 when the loss plateaus. $\lambda$ is set to 1 for our model. The random seed for all experiments are fixed as 42.
\vspace{-1em}

\subsection{Results}
\vspace{-0.5em}
The main results are shown in Table~\ref{tab:result}. As we can see from the last row of the table, our proposed method achieves the highest miAP on unseen classes in both datasets. We achieve the highest unseen mAP on MS-COCO and second highest unseen mAP on NUS-WIDE~\cite{chua2009nuswide}. Although our model has 1.64\% lower unseen mAP than visual-semantic~\cite{akata2013ALE} on NUS-WIDE~\cite{chua2009nuswide}, our unseen miAP is 7.77\% higher than visual-semantic~\cite{akata2013ALE}. For the MS-COCO~\cite{lin2014mscoco} dataset, our model performs better than   SKG~\cite{arjovsky2017wgan} with an obvious margin of about 3.5\% unseen miAP and 2.5\% unseen mAP. On NUS-WIDE~\cite{chua2009nuswide}, our RGCN-PosVAE outperforms the second best Fast0Tag~\cite{zhang2016fast} by around 2\% unseen miAP and 0.6\% unseen mAP.

Fast0tag~\cite{zhang2016fast} achieves higher seen and unseen miAP on both datasets than visual-semantic~\cite{akata2013ALE}, while visual-semantic~\cite{akata2013ALE} performs better on seen mAP than Fast0tag~\cite{zhang2016fast} on both datasets.
SKG~\cite{lee2018multi} performs better than the previous two baselines on MS-COCO~\cite{lin2014mscoco}, but is slightly worse than Fast0tag~\cite{zhang2016fast} on the NUS-WIDE~\cite{chua2009nuswide}. The RGCN baseline, which is the same as our full model but without incorporating ImageNet nodes into the graph, still outperforms Fast0Tag~\cite{zhang2016fast} on MS-COCO, and have comparable result with SKG~\cite{lee2018multi} on NUS-WIDE. By comparing the seen and unseen class performance of each model, we can see that other baselines generally achieve better performance on seen classes than our proposed model, but have lower performance on unseen classes, which indicates that our proposed model is less likely to overfit seen classes and better able to generalize to unseen classes than  others.

The basic RGCN model achieves higher unseen mAP than SKG~\cite{lee2018multi} on MS-COCO and slightly lower unseen miAP on NUS-WIDE~\cite{chua2009nuswide}. But their results are still comparable, which shows that our proposed RGCN itself is already a very strong baseline. RGCN-XL, which is an extended RGCN with ImageNet nodes, achieves higher unseen  miAP than RGCN, which shows the potential of utilizing the semantic information of ImageNet classes, although in a simple fashion. It can also be noted that RGCN-XL has much lower performance on seen classes on MS-COCO. The reason may be that the brute-force way of incorporating ImageNet classes will make the predictions on seen classes harder, since the input images do not actually have those ImageNet classes while the model is trying to treat them the same as the classes in the target dataset when initializing their node states. Meanwhile, the proposed RGCN-PosVAE model, which infers the initial states of ImageNet classes using variational inference, achieves higher scores on both seen and unseen classes than RGCN-XL, which demonstrates the advantages of using the proposed method to initialize the hidden states of ImageNet classes.
\vspace{-1.5em}
\subsection{Ablation Study}
\vspace{-1.5em}

\begin{figure}[!ht]
\centering
\subfigure[]{
\begin{minipage}[l]{0.36\linewidth}
\centering
\includegraphics[width=1\textwidth]{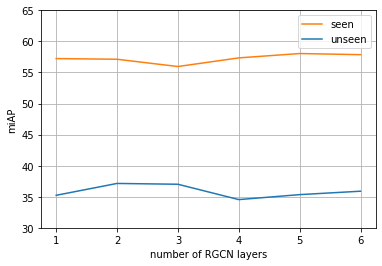}
\end{minipage}
\label{fig:layers}
}
\subfigure[]{
\begin{minipage}[l]{0.4\linewidth}
\centering
\includegraphics[width=1\textwidth]{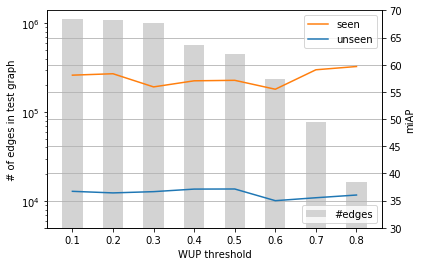}
\end{minipage}
\label{fig:threshold}
}
\vspace{-0.5em}
\caption{Ablation study}
\vspace{-1em}
\label{fig:ablation}
\end{figure}

\textbf{Effect of the number of GCN layers.} We first investigate how the number of RGCN layers affects our model's performance on the MS-COCO dataset~\cite{lin2014mscoco}. We try different numbers of layers from 1 up to 6, and plot the seen and unseen classes miAP in Figure~\ref{fig:layers}. As we can see, the miAP performance on seen classes does not vary a lot with different numbers of RGCN layers. On the other hand, two or three layers of RGCN have similar performance on unseen classes, while the version with only one layer has miAP about 2\% lower than the highest. After increasing the number of layers to 4, 5 and 6, the model's performance on unseen classes drops, which indicates that the model starts to become more overfitted to seen classes.

\textbf{Effect of the WUP similarity threshold.} We also investigate how the threshold of WUP similarity can affect our model's performance. We tune the threshold from 0.1 to 0.8, and show the result as well as the number of edges generated by each threshold in Figure~\ref{fig:threshold}. As we can see, although setting a low threshold will include some noisy edges that may be misleading, our model still achieves a good performance on unseen miAP, with less than 1\% decrease from the best one. As the threshold becomes larger than 0.5, we can see that there is an obvious drop of performance on unseen miAP. This is because the number of edges is much smaller when the threshold is too large, and thus the model cannot make good reference to seen classes when predicting unseen classes. Overall, our model's performance on seen and unseen classes is still relatively robust with respect to the WUP~\cite{wu1994wup} similarity threshold when the threshold is less than 0.6.
\vspace{-2em}



\section{Conclusion}
\vspace{-1em}
In this paper, we tackle the multi-label zero-shot learning (ML-ZSL) problem by first pointing out the potential semantic gap between seen and unseen classes, and propose to incorporate external ImageNet classes to help predict the seen can unseen classes. Specifically, we construct a knowledge graph consisting both clases from ImageNet and the target dataset (\eg, MS-COCO~\cite{lin2014mscoco} and NUS-WIDE~\cite{chua2009nuswide}), and design a PosVAE network to infer the node states of ImageNet classes, and learn a relational graph convolutional network (RGCN) which calculates the propagation weights between each pair of classes given their GloVe~\cite{pennington2014glove} embeddings. Experiments show that our proposed method has a clear advantage of predicting unseen classes.

\bibliography{reference}
\end{document}